\documentclass[sigconf]{acmart}
\renewcommand\footnotetextcopyrightpermission[1]{} 
\usepackage{url}
\usepackage{comment}
\usepackage{graphicx}
\usepackage{stfloats}
\usepackage{subfigure}

\AtBeginDocument{%
  \providecommand\BibTeX{{%
    \normalfont B\kern-0.5em{\scshape i\kern-0.25em b}\kern-0.8em\TeX}}}

\acmConference[EEKE 2020 @ JCDL '20] {1st Workshop on Extraction and Evaluation of Knowledge Entities from Scientific Documents}{August 1--5, 2020}{Virtual Event, China}
\acmBooktitle{ACM/IEEE Joint Conference on Digital Libraries in 2020 (JCDL '20), August 1--5, 2020, Virtual Event, China}

\begin{document}


\title{\textsc{NLPContributions}: An Annotation Scheme for \\ Machine Reading of Scholarly Contributions in \\ Natural Language Processing Literature}

\renewcommand{\shorttitle}{\textsc{NLPContributions}: An Annotation Scheme}

\author{Jennifer D'Souza}
\orcid{0000-0002-6616-9509}
\affiliation{%
  \institution{TIB Leibniz Information Centre for Science and Technology}
  \city{Hannover}
  \country{Germany}}
\email{jennifer.dsouza@tib.eu}

\author{S\"oren Auer}
\orcid{0000-0002-0698-2864}
\affiliation{%
  \institution{TIB Leibniz Information Centre for Science and Technology \& L3S Research Center}
  \city{Hannover}
  \country{Germany}}
\email{soeren.auer@tib.eu}

\renewcommand{\shortauthors}{D'Souza and Auer}

\begin{abstract}
We describe an annotation initiative to capture the scholarly contributions in natural language processing (NLP) articles, particularly, for the articles that discuss machine learning (ML) approaches for various information extraction tasks. We develop the annotation task based on a pilot annotation exercise on 50 NLP-ML scholarly articles presenting contributions to five information extraction tasks 1. machine translation, 2. named entity recognition, 3. question answering, 4. relation classification, and 5. text classification. In this article, we describe the outcomes of this pilot annotation phase. Through the exercise we have obtained an annotation methodology; and found ten core information units that reflect the contribution of the NLP-ML scholarly investigations. The resulting annotation scheme we developed based on these information units is called \textsc{NLPContributions}.

The overarching goal of our endeavor is four-fold: 1) to find a systematic set of patterns of subject-predicate-object statements for the semantic structuring of scholarly contributions that are more or less generically applicable for NLP-ML research articles; 2) to apply the discovered patterns in the creation of a larger annotated dataset for training machine readers~\cite{etzioni2006machine} of research contributions; 3) to ingest the dataset into the Open Research Knowledge Graph (ORKG) infrastructure as a showcase for creating user-friendly state-of-the-art overviews; 4) to integrate the machine readers into the ORKG to assist users in the manual curation of their respective article contributions. We envision that the \textsc{NLPContributions} methodology engenders a wider discussion on the topic toward its further refinement and development. Our pilot annotated dataset of 50 NLP-ML scholarly articles according to the \textsc{NLPContributions} scheme is openly available to the research community at \url{https://doi.org/10.25835/0019761}.

\end{abstract}



\maketitle

\section{Introduction}
As the rate of research publications increases~\cite{stm}, there is a growing need within digital libraries to equip researchers with alternative knowledge representations, other than the traditional document-based format, for keeping pace with the rapid research progress~\cite{auer_soren_2018}. In this regard, several efforts exist or are currently underway for semantifying scholarly articles for their improved machine interpretability and ease in comprehension~\cite{fathalla2017towards,Jaradeh2019ORKG,oelen2020generate,vogt2020}. These models equip experts with a tool for semantifying their scholarly publications ranging from strictly-ontologized methodologies~\cite{fathalla2017towards,vogt2020} to less-strict, flexible description schemes~\cite{Jaradeh2019ORKG,oelen2019comparing}, wherein the latter aim toward the bottom-up, data-driven discovery of an ontology. Consequently, knowledge graphs~\cite{ammar2018construction,auer2018towards} are being advocated as a promising alternative to the document-based format for representing scholarly knowledge for the enhanced content ingestion enabled via their fine-grained machine interpretability.

The automated semantic extraction from scholarly publications using text mining has seen early initiatives based on sentences as the basic unit of analysis. To this end, ontologies and vocabularies were created~\cite{teufel1999annotation,Soldatova2006AnOO,constantin2016document,pertsas2017scholarly}, corpora were annotated~\cite{Liakata2010CorporaFT,Fisas2016AMA}, and machine learning methods were applied~\cite{liakata2012automatic}. Recently, scientific IE has targeted search technology, thus newer corpora have been annotated at the phrasal unit of information with three or six types of scientific concepts in up to ten disciplines~\cite{handschuh2014acl,augenstein2017semeval,Luan2018MultiTaskIO,lrec2020} facilitating machine learning system development~\cite{Ammar2017TheAS,luan2017scientific,beltagy2019scibert,ecir2020}. In general, a phrase-focused annotation scheme more directly influences the building of a scholarly knowledge graph, since phrases constitute knowledge graph statements. Nonetheless, sentence-level annotations are just as poignant offering knowledge graph modelers better context from which the phrases are obtained for improved knowledge graph curation.

Over which, many recent data collection and annotation efforts~\cite{chemrecipes,labprotocols,mysore2019materials,kuniyoshi2020annotating} are steering new directions in text mining research on scholarly publications. These initiatives are focused on the shallow semantic structuring of the instructional content in lab protocols or descriptions of chemical synthesis reactions. This has entailed generating annotated datasets via structuring recipes to facilitate their automatic content mining for machine-actionable information which are presented otherwise in adhoc ways within scholarly documentation. Such datasets inadvertently facilitate the development of machine readers. In the past, such similar text mining research was conducted as the unsupervised mining of \textsc{Schemas} (also called \textit{scripts}, \textit{templates}, or \textit{frames})---as a generalization of recurring event knowledge (involving a sequence of three to ten events) with various participants~\cite{schank1977scripts}---primarily over newswire articles~\cite{chambers2008unsupervised,chambers2009unsupervised,balasubramanian2013generating,chambers2013event,simonson2015interactions,simonson2016nastea,simonson2018narrative}. They were a potent task at generalizing over similar but distinct narratives---can be seen as knowledge units---with the goal of revealing their underlying common elements. However, little insight was garnered on their practical task relevance. This has changed with the recent surface semantic structuring initiatives over instructional content. It has led to the realization of a seemingly new practicable direction that taps into the structuring of text and the structured information aggregation under \textsc{Scripts}-based knowledge themes. 

Since scientific literature is growing at a rapid rate and researchers today are faced with this publications deluge~\cite{jinha2010article}, it is increasingly tedious, if not practically impossible to keep up with the progress even within one's own narrow discipline. The Open Research Knowledge Graph (ORKG)~\cite{auer2018towards} is posited as a solution to the problem of keeping track of research progress minus the cognitive overload that reading dozens of full papers impose. It aims to build a comprehensive knowledge graph that publishes the research contributions of scholarly publications per paper, where the contributions are interconnected via the graph even across papers. At \url{https://www.orkg.org/} one can view the contribution knowledge graph of a single paper as a summary over its key contribution properties and values; or compare the contribution knowledge graphs over common properties across several papers in a tabulated survey. Practical examples of the latter can be found accessible online at \url{https://www.orkg.org/orkg/featured-comparisons}. This practically addresses the knowledge ingestion problem for researchers. How? With the ORKG comparisons feature, researchers are no longer faced with the daunting cognitive ingestion obstacle from manually scouring through dozens of papers of unstructured content in their field. Where this process traditionally would take several days or months, using the ORKG contributions comparison tabulated view, the task is reduced to just a few minutes. Assuming the individual paper contributions are structured in the ORKG, they can then simply deconstruct the graph, tap into the aspects they are interested in, and can enhance it for their purposes. Further, they can select multiple such paper graphs and with the click of a button generate their tabulated comparison. For additional details on systems and methods beyond just the contribution highlights, they can still choose to read the original articles, but this time around equipped with a better selective understanding of which articles they should read in depth. Of-course scholarly article abstracts are intended for this purpose, but they are not machine interpretable, in other words, they cannot be comparatively organized. Further, the unstructured abstracts representation still treats research as data silos, thus with this model, research endeavors, in general, continue to be susceptible to redundancy~\cite{ioannidis2016mass}, lacking a meaningful way of connecting structured and unstructured information.

\subsection{Our Contribution}


In this paper, we propose a surface semantically structured dataset of 50 scholarly articles for their research contributions in the field of natural language processing focused on machine learning applications (the NLP-ML domain) across five different information extraction tasks to be integrable within the ORKG. To this end, we (1) identify sentences in scholarly articles that reflect research contributions; (2) create structured (subject,predicate,object) annotations from these sentences by identifying mentions of the contribution candidate term phrases and their relations; and (3) group collections of such triples, that arise from either consecutive or non-consecutive sentences, under one of ten core information units that capture an aspect of the contribution of NLP-ML scholarly articles. These core information units are conceptually posited as thematic scripts~\cite{schank1977scripts}. The resulting model formalized from the pilot annotation exercise we call the \textsc{NLPContributions} scheme.

It has the following characteristics: (1) via a contribution-centered model, it makes realistic the otherwise forbidding task of semantically structuring full-text scholarly articles---our task only needs a surface structuring of the highlights of the approach which often can be found in the Title, the Abstract, one or two paragraphs in the Introduction, and in the Results section; (2) it offers guidance for a structuring methodology, albeit still encompassing subjective decisions to a certain degree, but overall presenting a uniform model for identifying and structuring contributions---note that without a model, such structuring decisions may not end up being comparable across users and their modeled papers (see Figure~\ref{fig:fig1}); (3) the dataset is annotated in JSON format since it preserves relation hierarchies; (4) the annotated data we produce can be practically leveraged within frameworks such as the ORKG that support structured scholarly content-based knowledge ingestion. With the integration of our semantically structured scholarly contributions data in the ORKG, we aim to address the tedious and time-consuming scholarly knowledge ingestion problem via its contributions comparison feature. And further, by using the graph-based model, we also address the problem of scholarly information produced as data silos, as the ORKG connects the structured information across papers.

\section{Background and Related Work}

\paragraph{\textbf{Sentence-based Annotations of Scholarly Publications}}

Early initiatives in semantically structuring scholarly publications focused on sentences as the basic unit of analysis. In these sentence-based annotation schemes, all annotation methodologies~\cite{teufel1999annotation,teufel2009towards,Liakata2010CorporaFT,Fisas2016AMA} have had very specific aims for scientific knowledge capture. Seminal works in this direction consider the CoreSC (Core Scientific Concepts) sentence-based annotation scheme~\cite{Liakata2010CorporaFT}. This scheme aimed to model in finer granularity, i.e. at the sentence-level, concepts that are necessary for the description of a scientific investigation, while traditional approaches employ section names serving as coarse-grained paragraph-level annotations. Such semantified scientific knowledge capture was apt at highlighting selected sentences within computer-based readers. In this application context, mere sectional information organization for papers was considered as missing the finer rhetorical semantic classifications. E.g., in a \textsc{Results} section, the author may also provide some sentences of \textit{background} information, which in a sentence-wise semantic labeling are called \textsc{Background} and not \textsc{Results}. As another sentence-based scheme is the Argument Zoning (AZ) scheme~\cite{teufel2009towards}. This scheme aimed at modeling the rhetorics around knowledge claims between the current work and cited work. They used semantic classes as ``Own\_Method,'' ``Own\_Result,'' ``Other,'' ``Previous\_Own,'' ``Aim,'' etc., each elaborating on the rhetorical path to various knowledge claims. This latter scheme was apt for citation summaries, sentiment analysis and the extraction of information pertaining to knowledge claims. In general, such complementary aims for the sentence-based semantification of scholarly publications can be fused to generate more comprehensive summaries.

\paragraph{\textbf{Phrase-based Annotations of Scholarly Publications}}

The trend towards scientific terminology mining methods in NLP steered the release of phrase-based annotated datasets in various domains. An early dataset in this line of work was the ACL RD-TEC corpus~\cite{handschuh2014acl} which identified seven conceptual classes for terms in the full-text of scholarly publications in Computational Linguistics, viz. \textit{Technology and Method}; \textit{Tool and Library}; \textit{Language Resource}; \textit{Language Resource Product}; \textit{Models}; \textit{Measures and Measurements}; and \textit{Other}. Similar to terminology mining is the task of scientific keyphrase extraction. Extracting keyphrases is an important task in publishing platforms as they help recommend articles to readers, highlight missing citations to authors, identify potential reviewers for submissions, and analyse research trends over time. Scientific keyphrases, in particular, of type \textit{Processes}, \textit{Tasks} and \textit{Materials} were the focus of the SemEval17 corpus annotations~\cite{augenstein2017semeval}. The dataset comprised annotations of the full text articles in Computer Science, Material Sciences, and Physics. Following suit was the SciERC corpus~\cite{Luan2018MultiTaskIO} of annotated abstracts from the Artificial Intelligence domain. It included annotations for six concepts, viz. \textit{Task}, \textit{Method}, \textit{Metric}, \textit{Material}, \textit{Other-Scientific Term}, and \textit{Generic}. Finally, in the realm of corpora having phrase-based annotations, was the recently introduced STEM-ECR corpus~\cite{lrec2020} notable for its multidisciplinarity including the Science, Technology, Engineering, and Medicine domains. It was annotated with four generic concept types, viz. \textit{Process}, \textit{Method}, \textit{Material}, and \textit{Data} that mapped across all domains, and further with terms grounded in the real-world via Wikipedia/Wiktionary links.

Next, we discuss related works that semantically model instructional scientific content. In these works, the overarching scientific knowledge capture theme is the end-to-end semantification of an experimental process.

\paragraph{\textbf{Shallow Semantic Structural Annotations of Instructional Content in Scholarly Publications}}

Increasingly, text mining initiatives are seeking out recipes or formulaic semantic patterns to automatically mine machine-actionable information from scholarly articles~\cite{chemrecipes,labprotocols,mysore2019materials,kuniyoshi2020annotating}.

In~\cite{labprotocols}, they annotate wet lab protocols, covering a large spectrum of experimental biology, including neurology, epigenetics, metabolomics, cancer and stem cell biology, with actions corresponding to lab procedures and their attributes including materials, instruments and devices used to perform specific actions. Thereby the protocols then constituted a prespecified machine-readable format as opposed to the ad-hoc documentation norm. Kulkarni et al.~\cite{labprotocols} even release a large human-annotated corpus of semantified wet lab protocols to facilitate machine learning of such shallow semantic parsing over natural language instructions. Within scholarly articles, such instructions are typically published in the Materials and Method section in Biology and Chemistry fields. 

Along similar lines, inorganic materials synthesis reactions and procedures continue to reside as natural language descriptions in the text of journal articles. There is a growing impetus in such fields to find ways to systematically reduce the time and effort required to synthesize novel materials that presently remains one of the grand challenges in the field. In~\cite{chemrecipes,mysore2019materials}, to facilitate machine learning models for automatic extraction of materials syntheses from text, they present datasets of synthesis procedures annotated with semantic structure by domain experts in Materials Science. The types of information captured include synthesis operations (i.e. predicates), and the materials, conditions, apparatus and other entities participating in each synthesis step.

The \textsc{NLPContributions} annotation methodology proposed in this paper draws on each of the earlier categorizations of related work. \textbf{First}, the full-text of scholarly articles including the Title and the Abstract are annotated in a sentence-wise granularity with the aim of the annotated sentences being only those restricted to the contributions of the investigation. We selectively consider the full-text of the article by focusing only on specific sections of the article such as the Abstract, Introduction, and the Results sections. Sometimes we also model the contribution highlights from the Approach/System description in case if the Introduction does not contain such pertinent information of the proposed model. We skip the Background, Related Work, and Conclusion sections altogether. These sentences are then grouped under one of ten main information units, viz. \textsc{ResearchProblem}, \textsc{Objective}, \textsc{Approach}, \textsc{Tasks}, \textsc{ExperimentalSetup}, \textsc{Hyperparameters}, \textsc{Baselines}, \textsc{Results}, and \textsc{AblationAnalysis}. Each of these units are defined in detail in the next section. \textbf{Second}, from the grouped contribution-centered sentences, we perform phrase-based annotations for (subject, predicate, object) triples to model in a knowledge graph. And \textbf{Third}, the resulting dataset has an overarching knowledge capture objective: capturing the contribution of the scholarly article and, in particular, to facilitate the training of machine readers for the purpose along the lines of the machine-interpretable wet-lab protocols.

\section{The NLPContributions Model}

\subsection{Goals}

The development of the \textsc{NLPContributions} annotation model was backed by four primary goals:
\begin{enumerate}
    \item We aim to produce a semantic representation based on existing work, that can be well motivated as an annotation scheme for the application domain of NLP-ML scholarly articles, and is specifically aimed at the knowledge capture of the \textit{contributions} in scholarly articles; 
    \item  The annotated scholarly contributions based on \textsc{NLPContributions} should be integrable in the Open Research Knowledge Graph (ORKG)\footnote{\url{https://www.orkg.org/orkg/}}--the state-of-the-art content-based knowledge capturing platform of scholarly articles' contributions. 
    \item The \textsc{NLPContributions} model should be useful to produce data for the development of machine learning models in the form of machine readers~\cite{etzioni2006machine} of scholarly contributions. Such trained models can serve to automatically extract such structured information for downstream applications, either in completely automated or semi-automated workflows as recommenders.\footnote{In future work, we will expand our current pilot annotated dataset of 50 articles with at least 400 additional similarly annotated articles to facilitate machine learning.}
    \item The \textsc{NLPContributions} model should be amenable to feedback via a consensus approval or content annotation change suggestions from a large group of authors toward their scholarly article contribution descriptions (an experiment that is beyond the scope of the present work and planned as following work).
\end{enumerate}

The \textsc{NLPContributions} annotation model is designed for building a knowledge graph. It is not ontologized, therefore, we assume a bottom-up data-driven design toward ontology discovery as more annotated contributions data is available. Nonetheless, we do propose a core skeleton model for organizing the information at the top-level KG nodes. This involves a root node called \textsc{Contribution}, following which, at the first level of the knowledge graph, are ten nodes representing core information units under which the scholarly contributions data is organized. 

\subsection{The Ten Core Information Units}

In this section, we describe the ten information units in our model.

\paragraph{\bf \textsc{ResearchProblem}} Determines the research challenge addressed by a contribution using the predicate \textit{hasResearchProblem}. By definition, it is the focus of the research investigation, in other words, \textit{the issue for which the solution must be obtained.}

The task entails identifying only the research problem addressed in the paper and not research problems in general. For instance, in the paper about the BioBERT word embeddings~\cite{lee2020biobert}, their research problem is just the `domain-customization of BERT' and not `biomedical text mining,' since it is a secondary objective.

The \textsc{ResearchProblem} is typically found in an article's Title, Abstract and first few paragraphs of the Introduction. The task involves annotating one or more sentences and precisely the research problem phrase boundaries in the sentences. 

The subsequent seven information objects are connected to \textsc{Contribution} via the generic predicate \textit{has}.

\paragraph{\bf \textsc{Approach}} Depending on the paper's content, is referred to as \textsc{Model} or \textsc{Method} or \textsc{Architecture} or \textsc{System} or \textsc{Application}. Essentially, this is the contribution of the paper as \textit{the solution proposed for the research problem}.

\begin{figure}[!tb]
\includegraphics[height=6.8cm]{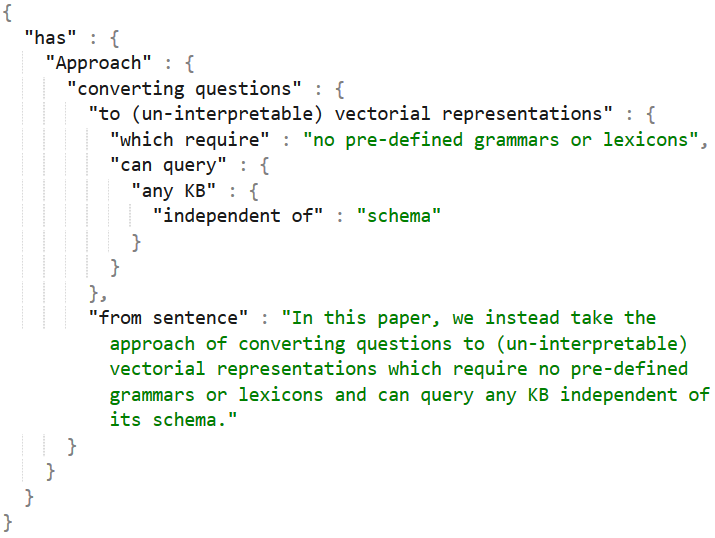}
\caption{Fine-grained modeling illustration from a single sentence for part of an \textsc{Approach} proposed in~\cite{bordes2014open}.}
\label{fig:eg1}
\end{figure}

The annotations are made only for the high-level overview of the approach without going into system details. Therefore, the equations associated with the model and all the system architecture figures are not part of the annotations. While annotating the earlier \textsc{ResearchProblem} did not involve semantic annotation granularity beyond one level, annotating the \textsc{Approach} can. Sometimes the annotations (one or multi-layered) are created using the elements within a single sentence itself (see Figure~\ref{fig:eg1}); at other times, if they are multi-layered semantic annotations, they are formed by bridging two or more sentences based on their coreference relations. For the annotation element content itself, while, in general, the subject, predicate, and object phrases are obtained directly from the sentence text, at times the predicate phrases have to be introduced as generic terms such as ``has'' or ``on'' or ``has description'' wherein the latter predicate is used for including, as objects, longer text fragments within a finer annotation granularity to describe the top-level node. The actual type of approach is restricted to those sub-types stated in the beginning of the paragraph and is decided based on the the reference to the solution used by the authors or the solution description section name itself. If the reference to the solution or its section name is specific to the paper, such as `Joint model,' then we rename it to just `Model.' In general, any alternate namings of the solution, other than those mentioned earlier, including ``idea'', are normalized to ``Model.'' Finally, as machine learning solutions, they are often given names. E.g., the model BioBERT~\cite{lee2020biobert}, in which case we introduce the predicate `called,' as in (Method, called, BioBERT).

The \textsc{Approach} is found in the article's Introduction section in the context of cue phrases such as ``we take the \textit{approach},'' ``we propose the \textit{model},'' ``our system \textit{architecture},'' or ``the \textit{method} proposed in this paper.'' However, there are exceptions when the Introduction does not present an overview of the system, in which case we analyze the first few lines within the main system description content in the article. Also, if the paper refers to their system by ``method'' or ``application,'' this is normalized to \textsc{Approach} information unit. \textsc{System} or \textsc{Architecture} is \textsc{Model} information unit.

\paragraph{\bf \textsc{Objective}} This is the \textit{defined function for the machine learning algorithm to optimize over}.

In some cases, the \textsc{Approach} objective is a complex function. In such cases, it is isolated as a separate information object connected directly to the \textsc{Contribution}.

\paragraph{\bf \textsc{ExperimentalSetup}} Has the alternate name \textsc{Hyperparameters}. It includes details about the platform including both hardware (e.g., GPU) and software (e.g., Tensorflow library) for implementing the machine learning solution; and of variables, that determine the network structure (e.g., number of hidden units) and how the network is trained (e.g., learning rate), for tuning the software to the task objective.

Recent machine learning models are all neural based and such models have several associated variables such as hidden units, model regularization parameters, learning rate, word embedding dimensions, etc. Thus to offer users a glance at the contributed system, this aspect is included in \textsc{NLPContributions}. We only model the experimental setup that are expressed in a few sentences or that are concisely tabulated. There are cases when the experimental setup is not modeled at all within \textsc{NLPContributions}. E.g., for the complex ``machine translation'' models that involve many parameters. Thus, whether the experimental setup should be modeled or not, may appear as a subjective decision, however, over the course of several annotated articles becomes apparent especially when the annotator begins to recognize the simple sentences that describe the experimental setup.

The \textsc{ExperimentalSetup} unit is found in the sections called Experiment, Experimental Setup, Implementation, Hyperparameters, or Training.

\paragraph{\bf \textsc{Results}} Are the main findings or outcomes reported in the article for the \textsc{ResearchProblem}. 

\begin{figure}[!tb]
\includegraphics[height=5cm]{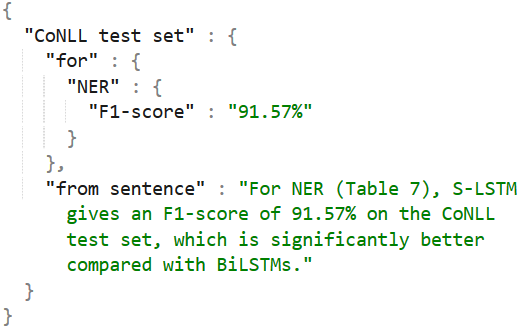}
\caption{Illustration of modeling of \textsc{Result} (from~\cite{zhang2018sentence}) w.r.t. a precedence of its elements as [dataset -> task -> metric -> score].}
\label{fig:eg2}
\end{figure}

Each \textsc{Result} unit involves some of the following elements: \{dataset, metric, task, performance score\}. Regardless of how the sentence(s) are written involving these elements, we assume the following precedence order: [dataset -> task -> metric -> score] or [task -> dataset -> metric -> score], as far as it can be applied without significantly changing the information in the sentence. Consider this illustrated in Figure~\ref{fig:eg2}. In the figure, the \textsc{JSON} is arranged starting at the dataset, followed by the task, then the metric, and finally the actual reported result. While this information unit is named per those stated in the earlier paragraph, if in a paper the section name is non-generic, e.g., ``Main results,'' ``End-to-end results,'' it is normalized to a default name ``Results.'' 

The \textsc{Results} unit is found in the Results, Experiments, or Tasks sections. While the results are often highlighted in the Introduction, unlike the \textsc{Approach} unit, in this case, we annotate the dedicated, detailed section on Results because results constitute a primary aspect of the contribution. Next we discuss the \textsc{Tasks} information unit, and note that \textsc{Results} can include \textsc{Tasks} and vice versa as we describe next.

\paragraph{\bf \textsc{Tasks}}: The \textsc{Approach} or \textsc{Model}, particularly in multi-task settings, are tested on more than one task, in which case, we list all the experimental tasks. The experimental tasks are often synonymous with the experimental datasets since it is common in NLP for tasks to be defined over datasets. Where lists of \textsc{Tasks} are concerned, the \textsc{Tasks} can include one or more of the \textsc{ExperimentalSetup}, \textsc{Hyperparameters}, and \textsc{Results} as sub information units. 

\paragraph{\bf \textsc{Experiments}} Are an encompassing information unit that includes one or more of the earlier discussed units. Can include a combination of \textsc{ExperimentalSetup} and \textsc{Results}, or it can be combination of lists of \textsc{Tasks} and their \textsc{Results}, or a combination of \textsc{Approach}, \textsc{ExperimentalSetup} and \textsc{Results}. 

Recently, more and more multitask systems are being developed. Consider, the BERT model~\cite{bert} as an example. Therefore, modeling \textsc{ExperimentalSetup} with \textsc{Results} or \textsc{Tasks} with \textsc{Results} is necessary in such systems since the experimental setup often changes per task producing a different set of results. Hence, this information unit encompassing two or more sub information units is relevant.

\paragraph{\bf \textsc{AblationAnalysis}} Is a form of \textsc{Results} that describes the performance of components in systems.

Unlike \textsc{Results}, \textsc{AblationAnalysis} is not performed in all papers. Further, in papers that have them, we only model these results if they are expressed in a few sentences, similar to our modeling decision for \textsc{Hyperparameters}.

The \textsc{AblationAnalysis} information unit is found in the sections that have Ablation in their title. Otherwise, it can also be found in the written text without having a dedicated section for it. For instance, in the paper ``End-to-End Relation Extraction using LSTMs on Sequences and Tree Structures''~\cite{miwa2016end} there is no section title with Ablation, but this information is extracted from the text via cue phrases that indicate ablation results are being discussed. 

\paragraph{\bf \textsc{Baselines}} are those listed systems that a proposed approach is compared against. 

The \textsc{Baselines} information unit is found in sections that have Baseline in their title. Otherwise, it can also be found in sections that are not directly titled Baseline, but require annotator judgement to infer that baseline systems are being discussed. For instance, in the paper ``Extracting Multiple-Relations in One-Pass with Pre-Trained Transformers,''~\cite{wang2019extracting} the baselines are discussed in subsection `Methods.' Or in paper ``Outrageously large neural networks: The sparsely-gated mixture-of-experts layer,''~\cite{shazeer2017outrageously}, the baselines are discussed in a section called ``Previous State-of-the-Art.''

Of these ten information units, only three are mandatory. They are \textsc{ResearchProblem}, \textsc{Approach}, and \textsc{Results}; the other seven may or may not be present depending on the content of the article.

\paragraph{\bf \textsc{Code}} is a link to the software on Github or on other similar open source platforms, or even on author's website.

\subsection{Contribution Sequences within Information Units}

Except for \textsc{ResearchProblem}, each of the remaining nine information units encapsulate different aspects of the contributions of scholarly investigations in the NLP-ML domain; with the \textsc{ResearchProblem} offering the primary contribution context. Within the seven different aspects, there are what we call Contribution Sequences. 

\begin{figure}[!tb]
\includegraphics[height=11cm]{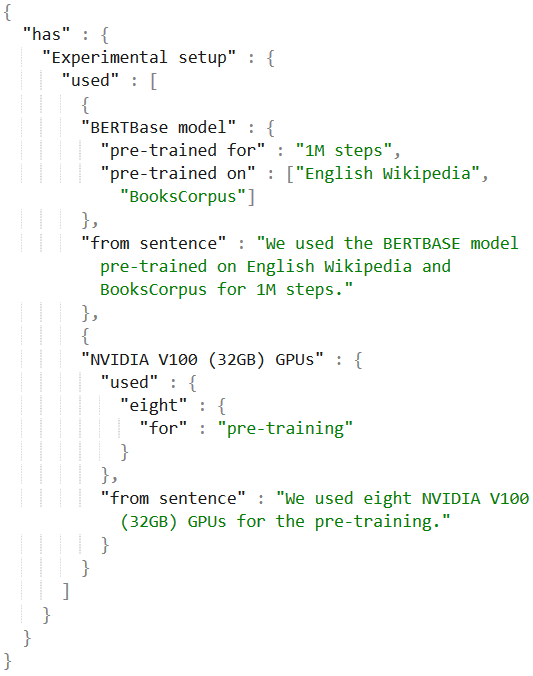}
\caption{Illustration of the modeling of Contribution Sequences in the \textsc{Experimental Setup} Information Unit (from~\cite{lee2020biobert}). \footnotesize{Created using \url{https://jsoneditoronline.org}}}
\label{fig:eg3}
\end{figure}

Here, with the help of an example depicted in Figure~\ref{fig:eg3} we illustrate the notion of contribution sequences. In this example, we model contribution sequences in the context of the \textsc{ExperimentalSetup} information unit. In the figure, this information unit has two contribution sequences. The first connected by predicate `used' to the object `BERTBase model,' and the second, also connected by predicate `used' to the object `NVIDIA V100 (32GB) GPUs.' The `BERTBase model' contribution sequence includes a second level of detail expressed via two different predicates `pre-trained for' and `pre-trained on.' As a model of scientific knowledge, the triple with the entities connected by the first predicate, i.e. (BERTBase model, pre-trained for, 1M steps) reflects that the `BertBase model' was pretrained for 1 million steps. The second predicate produces two triples: (BERTBase model, pre-trained on, English Wikipedia) and (BERTBase model, pre-trained on, BooksCorpus). In each case, the scientific knowledge captured by these two triples is that BERTBase was pretrained on \{Wikipedia, BooksCorpus\}. Note in the JSON data structure, the predicate connects the two objects as an array. Next, the second contribution sequence, hinged at `NVIDIA V100 (32GB) GPUs' as the subject has two levels of granularity. Consider the following three triples: (NVIDIA V100 (32GB) GPUs, used, ten) and (ten, for, pre-training). Note, in this nesting pattern, except for `NVIDIA V100 (32 GB) GPUs,' the predicates \{used, for\} and remaining entities \{ten, pre-training\} are nested according to their order of appearance in the written text. Therefore, in conclusion, an information unit can have several contribution sequences, and the contribution sequences need not be identically modeled. For instance, our second contribution sequence is modeled in a fine grained manner, i.e. in multiple levels. And when fine-grained modeling is employed, it is relatively straightforward to spot in the sentence(s) being modeled.

\section{The Pilot Annotation Task}

The pilot annotation task was performed by a postdoctoral researcher with a background in natural language processing. The \textsc{NLPContributions} model or scheme just described, were developed over the course of the pilot task. At a high-level, the annotations were performed in three main steps. They are presented next, after which we describe the annotation guidelines.

\subsection{Pilot Task Steps}

\noindent{\textbf{(a) Contribution-Focused Sentence Annotations.}} In this stage, sentences from scholarly articles were selected as candidate contribution sentences under each of the aforementioned mandatory three information units (viz., \textsc{ResearchProblem}, \textsc{Approach}, and \textsc{Results}) and, if applicable to the article, for one or more of the remaining seven information units as well.

To identify the contribution sentences in the article, the full-text of the article is searched. However, as discussed at the end of Section 2, the Background, Related Work, and Conclusions sections are entirely omitted from the search. Further, the section discussing the \textsc{Approach} or the \textsc{System} is only referred to when the Introduction section does not offer sufficient highlights of this information unit. In addition, except for tabulated hyperparameters, we do not consider other tables for annotation within the \textsc{NLPContributions} model.

To better clarify the pilot task process, in this subsection, we use Figure~\ref{fig:eg2} as the running example. From the example, at this stage, the sentence ``For NER (Table 7), S-LSTM gives an F1-score of 91.57\% on the CoNLL test set, which is significantly better compared with BiLSTMs.'' is selected as one of the contribution sentence candidates as part of the \textsc{Results} information unit. This sentence is selected from a Results subsection in ~\cite{zhang2018sentence}, but is just one among three others.

\noindent{\textbf{(b) Chunking Phrase Spans for Subject, Predicate, Object Entities.}} Then for the selected sentences, we annotate their scientific knowledge entities. The entities are annotated by annotators having an implicit understanding of whether they take the subject, predicate, or object roles in a per triple context. As a note, by our annotation scheme, predicates are not mandatorily verbs and can be nouns as well.

Resorting to our running example, for the selected sentence, this stage involves annotating the phrases ``For,'' ``NER,'' ``F1-score,'' ``91.57\%,'' and ``CoNLL test set,'' with the annotator cognizant of the fact that they will use the [dataset -> task -> metric -> score] scientific entity precedence in the next step.

\noindent{\textbf{(c) Creating contribution sequences.}} This involves relating the subjects and objects within triples, which as illustrated in Section 3.3, the object in one triple can be a subject in another triple if the annotation is performed at a fine-grained level of detail. For the most part, the nesting is done per order of appearance of the entities in the text, except for those involving the scientific entities \{dataset, task, metric, score\} under the \textsc{Results} information unit.

In the context of our running example, given the early annotated scientific entities, in this stage, the annotator will form the following two triples: (CoNLL test set, For, NER), (NER, F1-score, 91.57\%) as a single contribution sequence. What is not depicted in Figure 1 are the top-level annotations including the root node and one of the ten information unit nodes. This is modeled as follows: (Contribution, has, Results), and (Results, has, CoNLL test set).

\subsection{Task Guidelines}
In this section, we elicit a set of general guidelines that inform the annotation task.

\noindent{\textbf{\textit{How are information unit names selected?}}} For information units such as \textsc{Approach}, \textsc{ExperimentalSetup}, and \textsc{Results} that each have a set of candidate names, the applied name is the one selected based on the closest section title or cue phrase.

\noindent{\textbf{\textit{Which of the ten information units does the sentence belong to?}}} Conversely to the above, if a sentence is first identified as a contribution sentence candidate, it is placed within the information unit category that is identified directly based on the section header for the sentence in the paper or inferred from cue phrases from the first few sentences in its section.

\noindent{\textbf{\textit{Inferring Predicates.}}} In ideal settings, the constraint on the text used for subjects, objects, and predicates in contribution sequences is that they should be found in their corresponding sentence. However, for predicates this is not always possible. Since predicate information may not always be found in the text, it is sometimes annotated additionally based on the annotator judgment. However, even this open-ended choice remains restricted to a predefined set of candidates. It includes \{``has'', ``on'', ``by'', ``for'', ``has value'', ``has description'', ``based on'', ``called''\}. 


\noindent{\textbf{\textit{How are the supporting sentences linked to their corresponding contribution sequence within the overall JSON object?}}} The sentence(s) is stored in a dictionary with a ``from sentence'' key, which is then attached to either the first element or, if it is a nested triples hierarchy, sometimes even to the second element of a contribution sequence. The dictionary data-type containing the evidence sentence is either put as an array element, or as a nested dictionary element.

\noindent{\textbf{\textit{Are the nested contribution sequences always obtained from a single sentence?}}} The triples can be nested based on information from one or more sentences in the article. Further, the sentences need not be consecutive in the running text. As mentioned earlier, the evidence sentences are attached to the first element or the second element by the predicate ``from sentence.'' If a contribution sequence is generated from a table then the table number in the original paper is referenced.

\noindent{\textbf{\textit{When is the Approach actually modeled from the dedicated section as opposed to the Introduction?}}} In general, we avoid annotating the Approach or Model sections for their contribution sentences as they tend to delve deeply into the approach or model details, and involve complicated elements such as equations, etc. Instead, we restrict ourselves to the system higlights in the Introduction. However, in some articles the Introduction doesn't offer system highlights which is when we resort to using the dedicated section for the contribution highlights in this mandatory information unit.

\noindent{\textbf{\textit{Do we explore details about hardware used as part of the contribution?}}} Yes, if it is explicitly part of the hyperparameters. 

\noindent{\textbf{\textit{Are predicates always verbs?}}} Predicates are not always verbs. They can also be nouns especially in the hyperparameters section.

\noindent{\textbf{\textit{Creating contribution sequences from tabulated hyperparameters.}}} Only for hyperparameters, we model their tabulated version if given. This is done as follows: 1) for the predicate, we use the name of the parameter; and 2) for the object, the value against the name. Sometimes, however, if there are two-level hierarchical parameters, then the predicate is the first name, object is the value, and the value is qualified by the parameter name lower in the hierarchy. Qualifying the second name involves introducing the ``for'' predicate.

\begin{figure}[!tb]
\includegraphics[height=8.5cm]{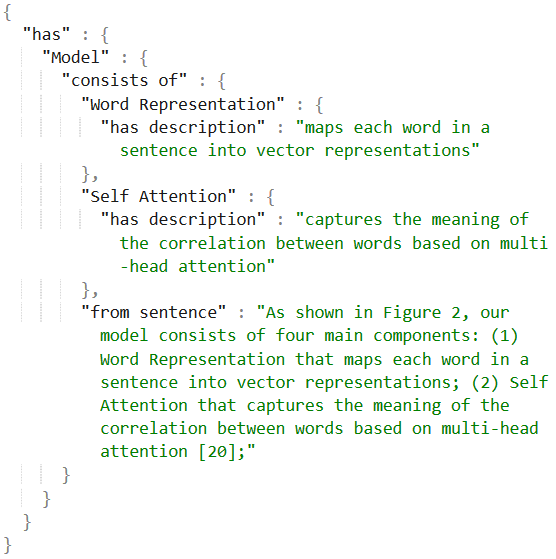}
\caption{Illustration of the modeling of a sentence with a list as part of the \textsc{Model} Information Unit (from~\cite{lee2019semantic}). \footnotesize{Created using \url{https://jsoneditoronline.org}}}
\label{fig:eg4}
\end{figure}

\noindent{\textbf{\textit{How are lists modeled within contribution sequences?}}} As part of the contribution sentence candidates, are also included sentences with lists. Such sentences are predominantly found for the \textsc{ExperimentalSetup} or \textsc{Result} information units. This is modeled as depicted in Figure~\ref{fig:eg4} for the first two list elements. Here, the \textsc{Model} information unit has two contribution sequences, each pertaining to a specific list item in the sentence. Further, the predicate ``has description'' is introduced for linking text descriptions.

\noindent{\textbf{\textit{Which JSON structures are used to represent the data?}}} Flexibly, they include dictionaries, or nested dictionaries, or arrays of items, where the items can be strings, dictionaries, nested dictionaries, or arrays themselves.

\noindent{\textbf{\textit{How are appositives handled?}}} We introduce a new predicate ``name'' to handle appositives.

\section{Materials and Tools}

\subsection{Paper Selection}

A collection of scholarly articles is downloaded based on the ones in the publicly available leaderboard of tasks in artificial intelligence called \url{https://paperswithcode.com/}. It predominantly represents papers in the Natural Language Processing and Computer Vision fields. For the purposes of our \textsc{NLPContributions} model, we restrict ourselves just to the NLP papers. From the set, we randomly select 10 papers in five different NLP-ML research tasks: 1. machine translation, 2. named entity recognition, 3. question answering, 4. relation classification, and 5. text classification.

\begin{figure}[!b]
\includegraphics[height=8cm]{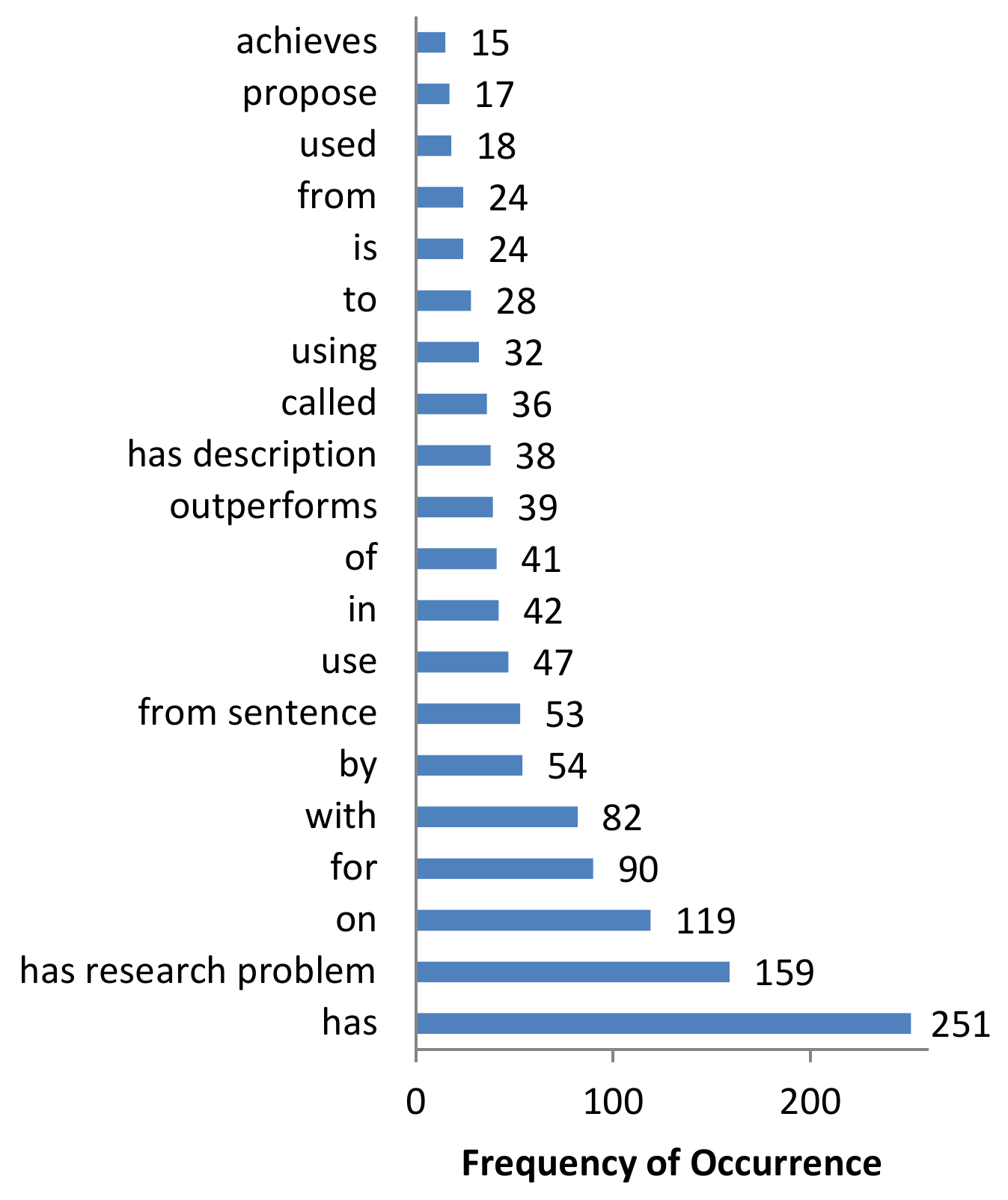}
\caption{A list of the predicates in our triples dataset that appear more than 15 times.}
\label{fig:predicates}
\end{figure}

\begin{figure*}
\subfigure[Research paper~\cite{soares2019matching} top-level snapshot in ORKG \url{https://www.orkg.org/orkg/paper/R41467/}]{\label{fig:a}\includegraphics[height=6cm]{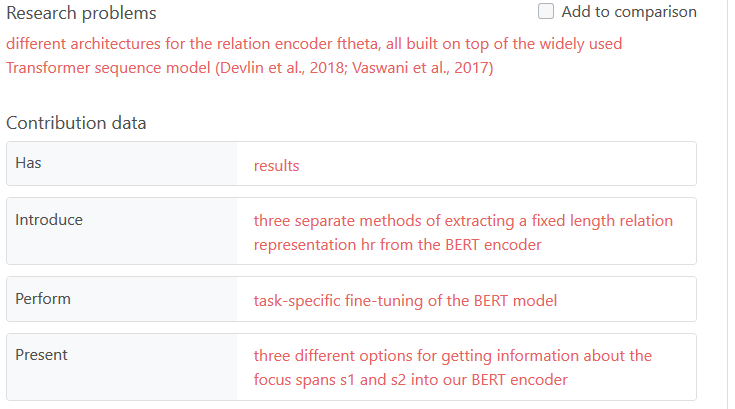}}
\subfigure[Research paper~\cite{guo2019attention} top-level snapshot in ORKG \url{https://www.orkg.org/orkg/paper/R41374}]{\label{fig:b}\includegraphics[height=7.5cm]{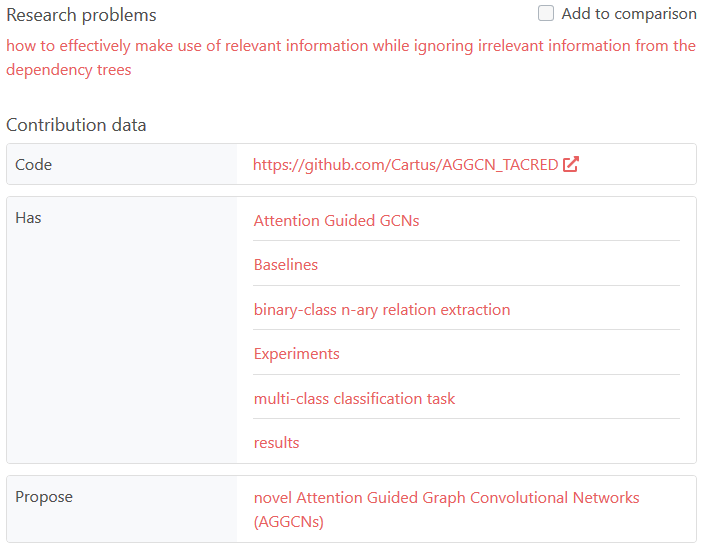}}
\subfigure[Research paper~\cite{zhang2018graph} top-level snapshot in ORKG \url{https://www.orkg.org/orkg/paper/R44287}]{\label{fig:c}\includegraphics[height=5cm]{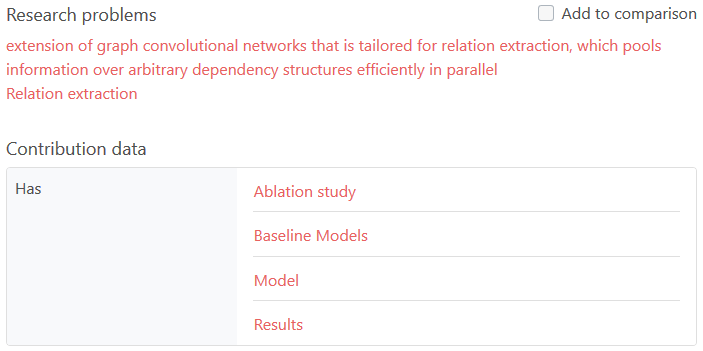}}
\caption{Figures~\ref{fig:a},\ref{fig:b},\ref{fig:c} depict evolution of the annotation scheme over three different research papers. Fig.~\ref{fig:c} is the resulting selected format \textsc{NLPContributions} that is proposed in this paper.}
\label{fig:fig1}
\end{figure*}

\subsection{Data Representation Format and Annotation Tools}

\textsc{JSON} was the chosen data format for storing the semantified parts of the scholarly articles contributions. To avoid syntax errors in creating the \textsc{JSON} objects, the annotations were made via \url{https://jsoneditoronline.org} which imposes valid \textsc{JSON} syntax checks. Finally, in the early stages of the annotation task, some of the annotations were made manually in the ORKG infrastructure~\url{https://www.orkg.org/orkg/} to test their practical suitability in a knowledge graph; three of such annotated papers are depicted in Figure~\ref{fig:fig1}. The links in the Figure captions can be visited to explore the annotations at their finer granularity of detail.

\subsection{Annotated Dataset Characteristics}

Overall, the annotated corpus contains a total of 2631 triples (avg. of 52 triples per article). Its data elements comprise 1033 unique subjects, 843 unique predicates, and 2182 unique objects. In Table~\ref{triple-dist} below, we show the per-task distribution of triples and their elements. Of all tasks, relation classification has the highest number of unique triples (544) and named entity recognition the least (473).

Generally, in the context of triples formation, predicates are often selected from a closed set and hence comprise a smaller group of items. In the \textsc{NLPContribution} model, however, predicates are extracted from the text if present. This leads to a much larger set of predicates that would require the application of predicate normalization functions to find the smaller core semantic set. In Figure~\ref{fig:predicates}, to offer some insights to this end, we show the predicates that appear more than 15 times over all the triples. We find the predicate \textit{has} appears most frequently since its function often serves as a filler predicate. A complete list of the predicates is released in our dataset repository online \url{ https://doi.org/10.25835/0019761}.

\begin{table}[!ht]
\begin{tabular}{l|l|l|l|l|l|}
          & MT  & NER & QA  & RC  & TC  \\ \hline
\textit{Subject}   & 259 & 209 & 203 & 228 & 221 \\
\textit{Predicate} & 243 & 220 & 187 & 201 & 252 \\
\textit{Object}    & 471 & 434 & 515 & 455 & 459 \\ \hline
Total     & 502 & 473 & 497 & 544 & 504
\end{tabular}
\caption{Per-task (machine translation (MT), named entity recognition (NER), question answering (QA), relation classification (RC), text classification (TC)) triples distribution in terms of unique subject, predicate, object, and overall.}
\label{triple-dist}
\end{table}

\section{Use Case: NLPContributions in ORKG}

As a use case of the ORKG infrastructure, instead of presenting just the annotations obtained from \textsc{NLPContributions}, we present a further enriched showcase. Specifically, we model the evolution of the annotation scheme at three different attempts with the third one arriving at \textsc{NLPContributions}. This is depicted in Figure~\ref{fig:fig1}. Our use case is an enriched one for two reasons: 1) it depicts the ORKG infrastructure flexibility for data-driven ontology discovery that makes allowances for different design decisions; and 2) it also shows how within flexible infrastructures the possibilities can be too wide that arriving at a consensus can potentially prove a challenge if it isn't mandated at a critical point in the data accumulation.

Figure~\ref{fig:a} depicts the first modeling attempts of an NLP-ML contribution. For predicates, the model restricts itself to use only those found in the text. The limitation of such a model is that not normalizing linguistic variations very rarely creates comparable models across investigations even if they imply the same thing. Hence, we found that for comparability a common predicate vocabulary at the top-level in the model minimally needs to be in place. Figure~\ref{fig:b} is the second attempt of modeling a different NLP-ML contribution. In this attempt, the predicates at the top-level are mostly normalized to a generic ``has,'' however, ``has'' is connected to various information items again lexically based on the text of the scholarly articles, one or more of which can be grouped under a common category. Via such observations, we systematized the knowledge organization at the top-level of the graph by introducing the ten information unit nodes. Figure~\ref{fig:c} is the resulting \textsc{NLPContributions} annotations model. Within this model, scholarly contributions with one or more of the information units in common, viz. ``Ablation study,'' ``Baseline Models,'' ``Model,'' and ``Results,'' can be uniformly compared.

\section{Limitations}

\noindent{\textbf{\textit{Obtaining disjoint (subject, predicate, object) triples as contribution sequences.}}} It was not possible to extract disjoint triples from all sentences. In many cases, we extract the main predicate and use as object the relevant full sentence or its clausal part. From~\cite{lee2020biobert}, for instance, under the \textsc{ExperimentalResults} information unit, we model the following: (Contribution, has, Experimental results); (Experimental results, on, all datasets); and (all datasets, achieves, BioBERT achieves higher scores than BERT). Note, in the last triple, ``achieves'' was used as a predicate and its object ``BioBERT achieves higher scores than BERT'' is modeled as a clausal sentence part.

\noindent{\textbf{\textit{Employing coreference relations between scientific entities.}}} In the fine-grained modeling of schemas, scientific entities within triples are sometimes nested across sentences by leveraging their coreference relations. We consider this a limitation toward the automated machine reading task, since coreference resolution itself is often challenging to perform automatically.

\noindent{\textbf{\textit{Tabulated results are not incorporated within \textsc{NLPContributions}.}}} Unlike tabulated hyperparameters which have a standard format, tabulated results have significantly varying formats. Thus their automated table parsing is a challenging task in itself. Nonetheless, by considering the textual results, we relegate ourselves to their summarized description, which often serves sufficient for highlighting the contribution.

\noindent{\textbf{\textit{Can all NLP-ML papers be modeled by \textsc{NLPContributions?}}}} While we can conclude that some papers are easier to model than others (e.g., articles addressing `relation extraction' vs. `machine translation' which are harder), it is possible that all papers can be modelled by at least some if not all the information units of the model we propose.



\section{Discussion}

From the pilot dataset annotation exercise, we note the following regarding task practically. Knowledge modeled under some information units are more amenable to systematic structuring than others. E.g., information units such as \textsc{ResearchProblem}, \textsc{ExperimentalSetup}, \textsc{Results}, and \textsc{Baselines} are readily amenable for systematic templates discovery toward their structured modeling within the ORKG; whereas the remaining information units, especially \textsc{Approach} or \textsc{Model}, will require additional normalization steps toward the search for their better structuring.

\section{Conclusions and Future Directions}

The Open Research Knowledge Graph~\cite{auer_soren_2018} makes scholarly knowledge about research contributions machine-actionable: i.e. findable, structured, and comparable. Manually building such a knowledge graph is time-consuming and requires the expertise of paper authors and domain experts. In order to efficiently build a scholarly knowledge contributions graph, we will leverage the technology of machine readers~\cite{etzioni2006machine} to assist the user in annotating scholarly article contributions. But the machine readers will need to be trained for such a task objective. To this end, in this work, we have proposed an annotation scheme for capturing the contributions in natural language processing scholarly articles, in order to create such training datasets for machine readers. In addition, we also provide a set of 50 annotated articles by the \textsc{NLPContributions} scheme as a practical demonstration of feasibility of the annotation task. However, for the training of machine learning models in future work we will release a larger dataset annotated by the proposed scheme. To facilitate future research, our pilot dataset is released online at \url{ https://doi.org/10.25835/0019761}.

Finally, aligned with the initiatives within research communities to build the Internet of FAIR Data and Services (IFDS)~\cite{ayris_2016}, the data within ORKG are compliant~\cite{oelen2020generate} with such FAIR data principles~\cite{wilkinson2016fair} thus making them Findable, Accessible, Interoperable and Reusable. Since the dataset we annotate by our proposed scheme is designed to be ORKG-compliant, we adopt the cutting-edge standard of data creation within the research community.

Nevertheless, the \textsc{NLPContribution} model is a surface semantic structuring scheme for the contributions in unstructured text. To realize a full-fledged machine-actionable and inferenceable knowledge graph of scholarly contributions, as future directions, there are a few IE modules that would need to be improved or added. They are (1) improving the PDF parser to produce less noisy output; (2) incorporating an entity and relation linking and normalization module; (3) merging phrases from the unstructured text with known ontologies (e.g., the MEX vocabulary~\cite{mex}) to align resources and thus ensure data interoperability and reusability; and (4) extending the model to more scholarly disciplines and domains.


\bibliographystyle{ACM-Reference-Format}
\bibliography{sample-sigconf}

\appendix

\section{Twice Modeling Agreement}

In general, even if the annotations are performed by a single annotator, there will be an annotation discrepancy. Compare the same information unit “Experimental Setup” modeled in Figure 7 below versus that modeled in Figure 3. Fig. 7 was the first annotation attempt and includes the second attempted model, done on a different day and blind from the the first. While neither are incorrect, the second
has taken the least annotated information route possibly due to
annotator fatigue, hence a two-pass methodology is recommended.

\begin{figure*}[b]
\centering
\includegraphics[height=17cm]{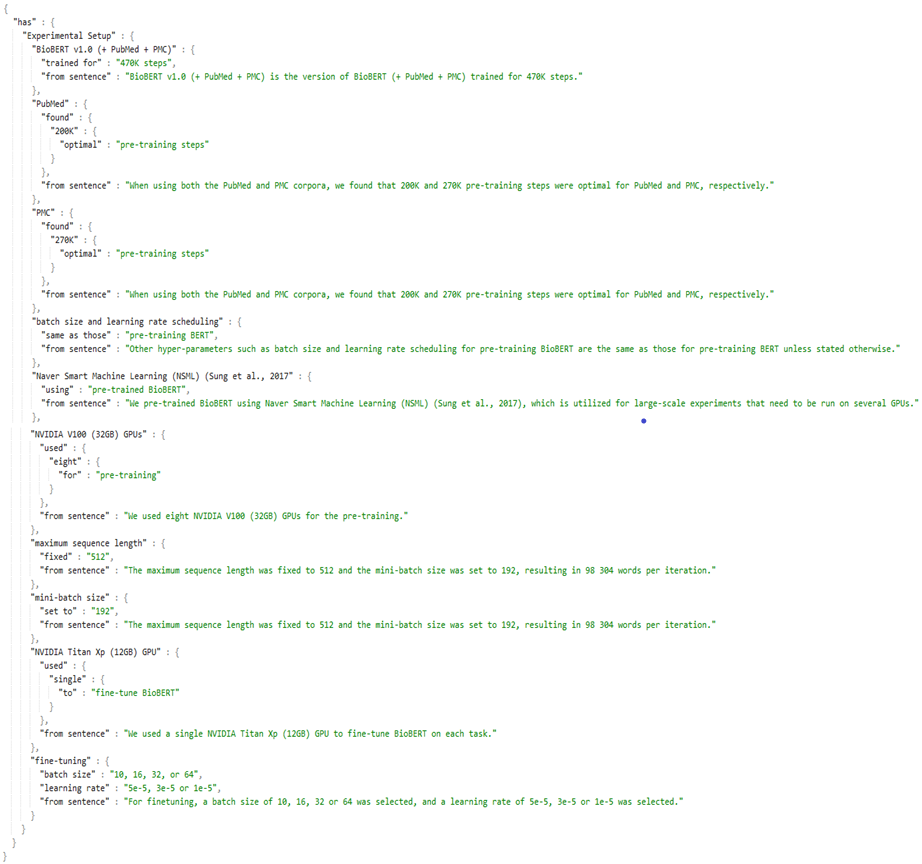}
\caption{Illustration of modeling of Contribution Sequences in the \textsc{Experimental Setup} Information Unit (from~\cite{lee2020biobert}) in a first annotation attempt. Contrast with second attempt depicted in Figure~\ref{fig:eg3} in the main paper content.}
\label{fig:eg3-det}
\end{figure*}




\end{document}